\documentclass[10pt,twocolumn,letterpaper]{article}
\usepackage{iccv}
\usepackage{times}
\usepackage{epsfig}
\usepackage{graphicx}
\usepackage{amsmath}
\usepackage{amssymb}
\usepackage{multirow}
\usepackage{tablefootnote}
\usepackage{threeparttable}
\usepackage[shortcuts]{extdash}

\usepackage[pagebackref=true,breaklinks=true,letterpaper=true,colorlinks,bookmarks=false]{hyperref}

\iccvfinalcopy 

\def\iccvPaperID{****} 

\begin{document}

\title{Adaptive Context Encoding Module for Semantic Segmentation}

\author{Congcong Wang{$^1$}\\
\and Faouzi Alaya Cheikh{$^1$}\\
\and Azeddine Beghdadi{$^2$}\\
\and  Ole Jakob Elle{$^{3,4}$}\\
{$^1$} Norwegian Colour and Visual Computing Lab, NTNU, Norway.\\
{$^2$} L2TI-Institut Galil\'{e}e, Universit\'{e} Paris 13, Sorbonne Paris Cit\'{e}, Villetaneuse, France.\\
{$^3$} The Intervention Centre, Oslo University Hospital, Oslo, Norway.\\
{$^4$} The Department of Informatics, University of Oslo, Oslo, Norway.\\
{\tt\small \{congcong.wang,faouzi.cheikh\}@ntnu.no, azeddine.beghdadi@univ-paris13.fr, oelle@ous-hf.no}
}

\maketitle

\begin{abstract}
The object sizes in images are diverse, therefore, capturing multiple scale context information is essential for semantic segmentation. Existing context aggregation methods such as pyramid pooling module (PPM) and atrous spatial pyramid pooling (ASPP)  design different pooling size or atrous rate, such that multiple scale information is captured. However, the pooling sizes and atrous rates are chosen  manually and
empirically. In order to capture object context information adaptively, in this paper, we propose an adaptive context encoding (ACE) module based on deformable convolution operation to argument multiple scale information. Our ACE module can be embedded into other  Convolutional Neural Networks (CNN)  easily for context aggregation. The effectiveness of the proposed module is demonstrated on Pascal-Context and ADE20K datasets. Although our proposed ACE only consists of three deformable convolution blocks, it outperforms PPM and ASPP in terms of mean Intersection of Union (mIoU) on both datasets. All the experiment study confirms that our proposed module is effective as compared to the state-of-the-art methods.
\end{abstract}

\section{Introduction}
\label{sec:intro}
Semantic segmentation is a pixel wise classification problem, where class prediction is assigned for each pixel of an image. The development of deep learning brings semantic segmentation into a new era. Starting from the Fully Covolutional Network (FCN)~\cite{long2015fully}, we have seen a rapid increase in the research field of semantic segmentation based on Convolutional Neural Networks (CNN)~\cite{ronneberger2015u,chen2018deeplab,badrinarayanan2017segnet}. Those methods boost the field and reach the state-of-the-art performance on several semantic segmentation benchmarks.

However, multiple scale objects context understanding is still a challenging problem. Some approaches have been proposed to handle this problem. Following similar criterion presented in~\cite{chen2017rethinking}, we group those methods into three categories. First, image pyramid based methods: the input image is decomposed to image pyramid, then DCNN (Deep Convolutional Neural Network) is applied separately to every resolution level of the image pyramid input~\cite{farabet2013learning, pinheiro2014recurrent, lin2016efficient}. In this way, different scale objects are captured from different level feature maps. Second, encoder-decoder based methods: for encoder, convolution and pooling operations are applied hierarchically to extract features, then the spatial resolution is recovered in the decoder path by hierarchically up-sampling and convolution operations. The most representable architecture is U-Net~\cite{ronneberger2015u} which has been achieving promising result in medical image processing field. And many other encoder-decoder based architectures ~\cite{badrinarayanan2017segnet,lin2017refinenet,amirul2017gated,mohammed2018net,mohammed2019streoscennet}. Third, spatial pyramid pooling strategy based approaches: the feature maps are aggregated by pooling operations or by atrous convolutions with multiple rates. The atrous spatial pyramid pooling (ASPP) proposed in DeepLabs~\cite{chen2018deeplab, chen2017rethinking} and the pyramid pooling module (PPM) presented in PSPNet ~\cite{zhao2017pyramid} are two representative work in this group.

The performance of Deeplabs~\cite{chen2017rethinking, chen2018encoder} and PSPNet~\cite{zhao2017pyramid} on some benchmarks shows the effectiveness of their pyramid pooling module. However, the rates of ASPP and PPM  are manually selected, which still cannot flexibly and image dependently encode multi\=/scale information. 
In this paper, our goal is to investigate is there a way to adaptively and input image dependently aggregate the feature maps?

Rethinking of ASPP and atrous convolution, setting different values of the atrous rate of atrous convolution operation endows the network with multiple effective field\=/of\=/view, thus the ability of capturing multi-scale context information. Therefore, if we can adaptively adjust the field\=/of\=/view of the convolution operations in the aggregation part, it would be possible for the network to aggregate contextual information input dependently.

Interestingly, Deformable Convolution Networks (DCN) is proposed recently in~\cite{dai2017deformable, zhu2018deformable}. For deformable convolution, the sampling locations are learnable, which can be highly integrated into our purpose. Therefore, in this paper, we propose an adaptive context encoding (ACE) module based on deformable convolution, more precisely, we replace the ASPP module or PPM  by three deformable convolution blocks. 

This idea is evaluated on Pascal-Context~\cite{oliva2007role} and ADE20K~\cite{zhou2017scene} datasets for semantic segmentation. We experimentally demonstrate that our proposed method improves the segmentation result consistently over the baseline methods: ASPP  and PPM. Especially, a more robust performance is shown under different bath size settings during training process.
Moreover, even though our goal is to find a better multi-scale aggregation module compared to ASPP and PPM, our method achieves the state-of-the-art on Pascal-Context dataset with 53.6\% mIoU  and promising results on ADE20K dataset with a final score of 0.5535. Furthermore, our  ACE module can be easily embedded into other networks for further improvement.


The reminder of this paper is organized as follows. First, in Sec.~\ref{sec:related_work}, we review the related work on semantic segmentation. Next, in Sec~\ref{sec:method}, we present our proposed method. Then in~Sec.~\ref{sec:experiment}, the experimental results and analysis are presented. Finally, discussion and conclusions are drawn in Sec.~\ref{sec:conclusion}.

\section{Related Work}
\label{sec:related_work}

Deep learning based semantic segmentation is rapidly developed and great progress has been achieved. DCNN with pooling and convolution with striding operation is invariant to local image transformations, thus can extract abstractions of data hierarchically~\cite{zeiler2014visualizing}. On one hand, this ability is beneficial for high-level vision tasks such as classification. On the other hand, it can bound the performance of pixel wise dense prediction tasks where spatial information is important~\cite{chen2014semantic}. Semantic segmentation thus is challenging as it needs to simultaneously perform classification and localization.

There are many works proposed to improve semantic segmentation which can be briefly divided into two directions: Resolution Enlarging and Context Extraction.
\subsection{Resolution Enlarging}
Atrous convolution which is inspired by the atrous algorithm~\cite{mallat1999wavelet} is claimed to be useful for extracting denser feature maps which can further alleviate the detail information loss. Thus it is widely used in semantic segmentation to enlarge the receptive-field-of-view and extract dense feature~\cite{chen2018deeplab,chen2018encoder,wu2019fastfcn}. 
Besides, encoder-decoder architectures employ decoder to up-sampling the resolution hierarchically and composite for the information loss of encoder~\cite{ronneberger2015u,lin2017refinenet,yu2018learning}. 
Thus, we briefly group them into two categories: atrous convolution and encoder-decoder based approaches~\cite{wu2019fastfcn,yu2018bisenet}, which will be introduced in the following paragraphs.

\textbf{Atrous convolution based methods} Typically, for DCNN, such as Resnet~\cite{he2016deep}, the spatial resolution of the output feature maps of the final layer is 32 times smaller compared to the resolution of input images, which is harmful for pixel wise tasks.  Atrous convolution is used to enlarge the receptive field while preserving the resolution of the feature map. DeepLabs especially Deeplabv2~\cite{chen2018deeplab} and Deeplabv3~\cite{chen2017rethinking} are a series of methods which investigate atrous convolution for semantic segmentation and are considered as one of the state-of-the-art techniques. Similar feature extraction DCNN backbone is also used in PSPNet~\cite{zhao2017pyramid}, where the resolution of the final layer feature maps is 8 times smaller. Atrous convolution is an effective solution for spatial information loss. However, the larger feature maps and larger convolution kernels make the network require higher computational resource. Recently, in~\cite{wu2019fastfcn}, Wu~\textit{et al.} propose Joint Pyramid Up-sampling (JPU) to reduce the memory and time consuming atrous convolutions, while keeping the ability of extracting high resolution feature maps.

\textbf{Encoder-Decoder based methods} In an encoder-decoder network, the spatial resolution is gradually up-sampled at the decoder part. DeconvNet~\cite{noh2015learning} uses deconvolutional layers~\cite{noh2015learning} to recover the resolution which can get full resolution final prediction by a complex decoder part. U-Net~\cite{ronneberger2015u} introduces skip connections from encoder to decoder, thus the information from the skip connection is used to compensate for the information loss. RefineNet~\cite{lin2017refinenet} elaborately design an up-sampling path to fuse low level and high level features. DeepLabv3+~\cite{chen2018encoder} employs both skip connection and atrous convolution, thus reaching the state-of-the-art performance on some benchmarks to date. 

\begin{figure*}[ht!]
\centering
	\begin{minipage}[b]{.49\linewidth}
		\centering
		\centerline{\includegraphics[width=1\linewidth]{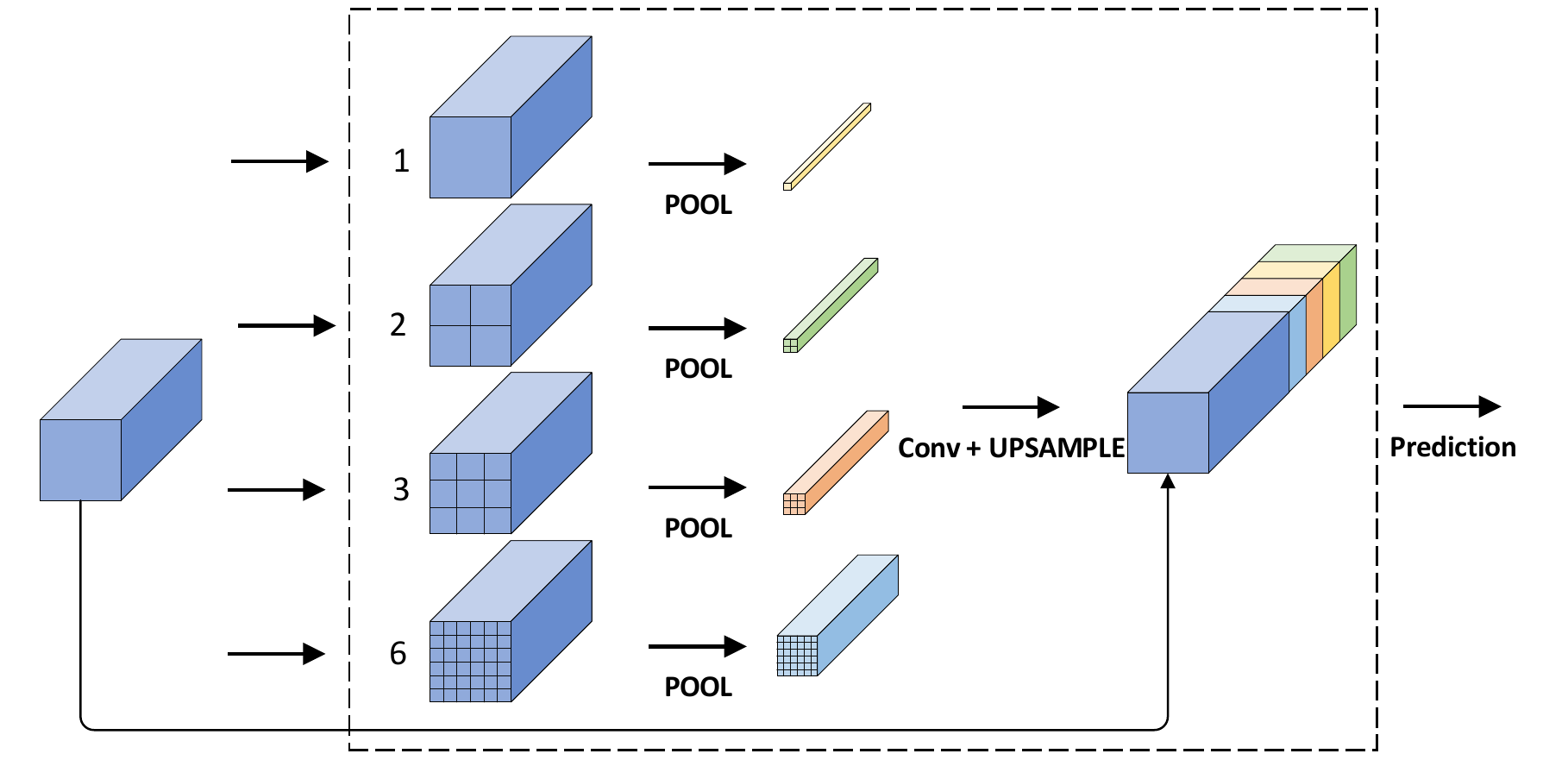}}
		\centerline{\small (a) PPM~\cite{zhao2017pyramid} }
	\end{minipage}
		\begin{minipage}[b]{.49\linewidth}
		\centering
		\centerline{\includegraphics[width=1\linewidth]{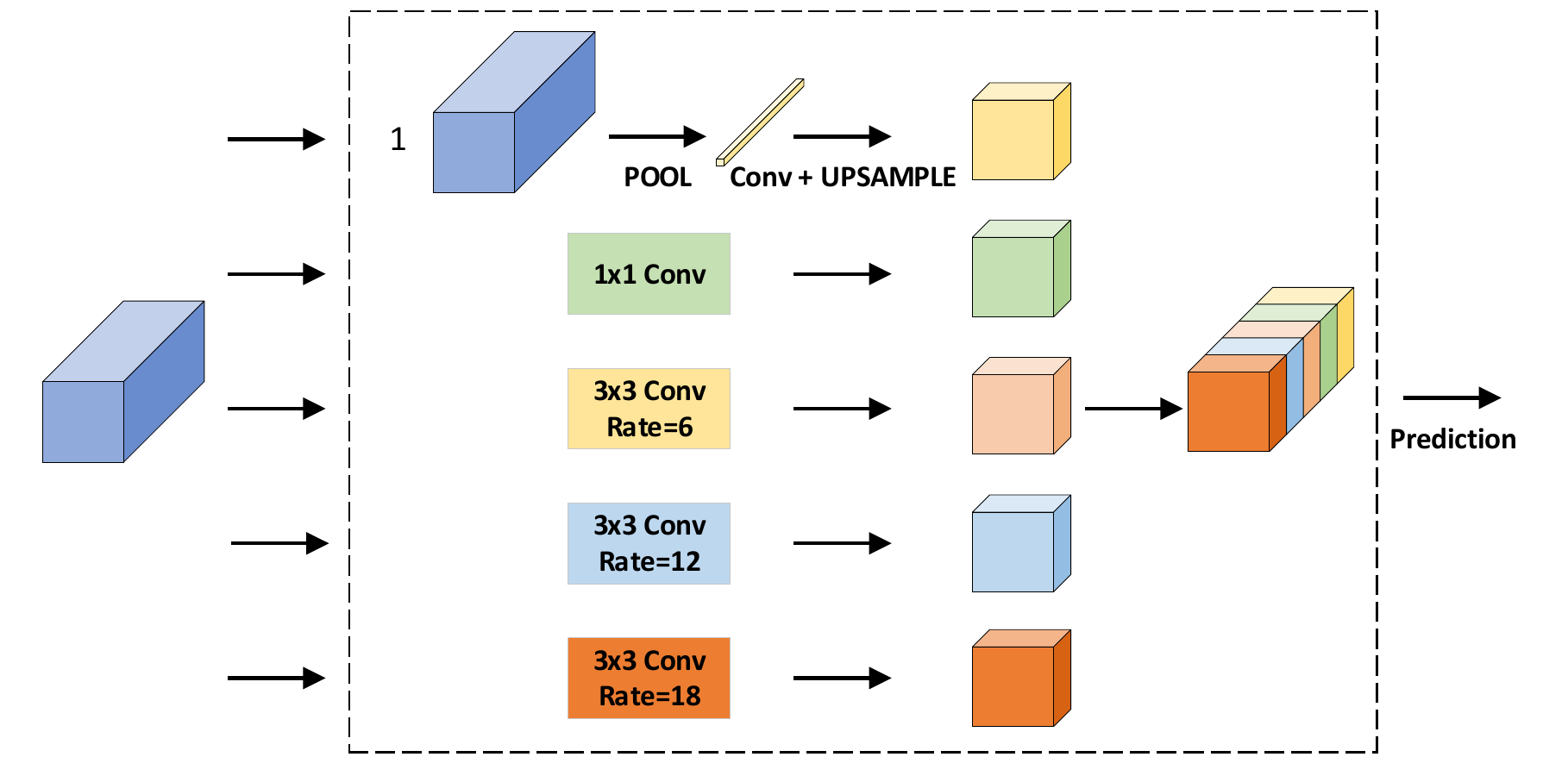}}
		\centerline{ \small (b) ASPP~\cite{chen2018deeplab}}
	\end{minipage}
    \vspace{0.15cm}
	\caption{ (a) Pyramid pooling module (PPM) proposed in PSPNet~\cite{zhao2017pyramid}. (b) Atrous spatial pyramid pooling (ASPP) module proposed in DeepLabs~\cite{chen2017rethinking,chen2018deeplab}.}
	\label{fig:ppmaspp}
\end{figure*}
\subsection{Context Extraction}
Scene context is important for extracting semantics. There are many approaches proposed to extract useful context information. Spatial pyramid pooling is proven to be effective for extracting context information~\cite{liu2015parsenet,zhao2017pyramid,chen2018deeplab}. Moreover, attention mechanism is proposed to learn the object context map in~\cite{yu2018learning}. In~\cite{peng2017large}, Peng~\textit{et al.} use convolution operations with large kernel size to extract classification information. Among those approaches, the spatial pyramid pooling based methods are popular.  Spatial pyramid pooling aims for extracting multiple scale context information from feature maps. For semantic segmentation, pyramid pooling module (PPM)~\cite{zhao2017pyramid} investigates pooling operation as a tool for multiple scale context aggregation and atrous spatial pyramid pooling (ASPP)~\cite{chen2017rethinking} exploits atrous convolution for pyramid pooling. 
These two modules   will be described in detail as it is highly related to the proposed approach.

\textbf{Pyramid Pooling Module (PPM)}
Global Average Pooling (GAP) is used to obtain global contextual prior in ParseNet~\cite{liu2015parsenet} for semantic segmentation. However, as pointed out in~\cite{zhao2017pyramid}, fusing one feature map into one single value may cause information loss. Thus, in~\cite{zhao2017pyramid}, Zhao~\textit{et al.} propose to hierarchically apply pooling operations with four scales as illustrated in Figure~\ref{fig:ppmaspp} (a), resulting in feature maps with four levels of resolution. The coarsest level is obtained by applying GAP on the feature maps and get a single vector output. For the other levels, the feature maps are first divided into sub-regions, then a global pooling is applied to every sub-region. The numbers of sub-region are set to $2\times2$, $3\times3$, $6\times6$ for each level respectively in paper~\cite{zhao2017pyramid} and illustrated in Figure~\ref{fig:ppmaspp} (a). This PPM  can thus extract the information at different scales for context aggregation.

\textbf{Atrous Spatial Pyramid Pooling (ASPP)}
ASPP module is first proposed in~\cite{chen2018deeplab} and further revised in~\cite{chen2017rethinking}. In ASPP module, as shown in Figure~\ref{fig:ppmaspp} (b), different atrous rates are used to extract multiple scale information.  Besides, in order to capture global context prior, similar to ParseNet~\cite{liu2015parsenet} and PPM~\cite{zhao2017pyramid}, GAP is applied. In summary,  one $1 \times 1$ convolution block and three $3 \times 3$ atrous convolution blocks with different atrous rates (6, 12, 18 respectively), and one GAP block are employed in parallel.

While Deeplabs and PSPNet reach the state-of-the-art performance on different benchmarks when they are proposed , and they are still have influence in semantic segmentation, it is important and interesting to investigate two following aspects: (1) It is obviously observed that the numbers of sub-region of PPM in PSPNet and the atrous rates of ASPP module from Deeplabs are selected empirically. The choice of those parameters need to be adjusted according to the applications, such as in~\cite{gu2019net}, $2\times2$, $3\times3$, $5\times5$, $6\times6$ sub-regions are chosen for their RMP (Residual Multi-kernel Pooling) module which is similar to PPM. It is essential to avoid choosing those parameters manually and empirically.
 (2)  PPM and ASPP both extract the context information by sampling from rigid rectangle regions which contain pixels from different object categories. However, for a certain pixel, the surrounding pixels which belong to the same category should contribute more. As also pointed out in~\cite{yuan2018ocnet}, Yuan and Wang hold similar opinion and they define object context as the set of pixels which belongs to the same category. While Yuan and Wang utilize self-attention~\cite{vaswani2017attention} mechanism to exploit context from pixels that from the same object class, in this work, inspired by the original ASPP, we will investigate the possibility to aggregate multi-scale information by adjusting adaptively the field-of-view of the convolution operation.

\section{Method}
\label{sec:method}
\begin{figure*}[ht!]
\centering
	\begin{minipage}[b]{0.9\linewidth}
		\centering
		\centerline{\includegraphics[width=1\linewidth]{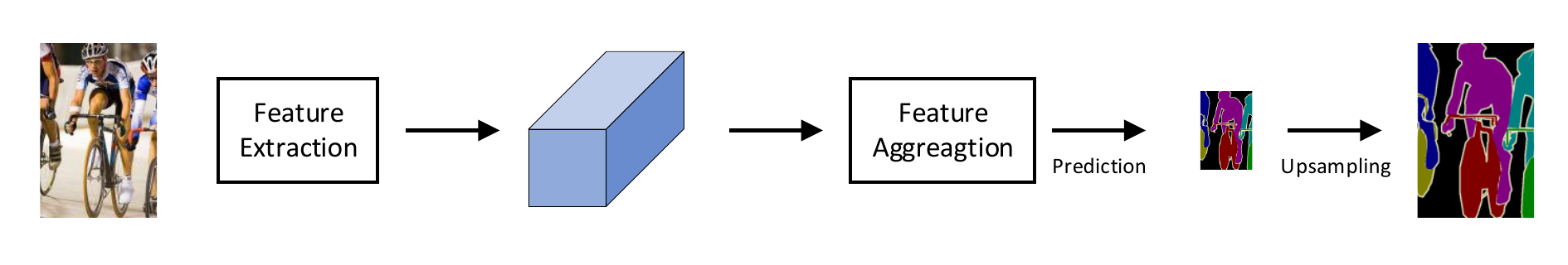}}
		\centerline{\small (a)}
	\end{minipage}
	\vspace{10pt}
	
		\begin{minipage}[b]{.75\linewidth}
		\centering
		\centerline{\includegraphics[width=1\linewidth]{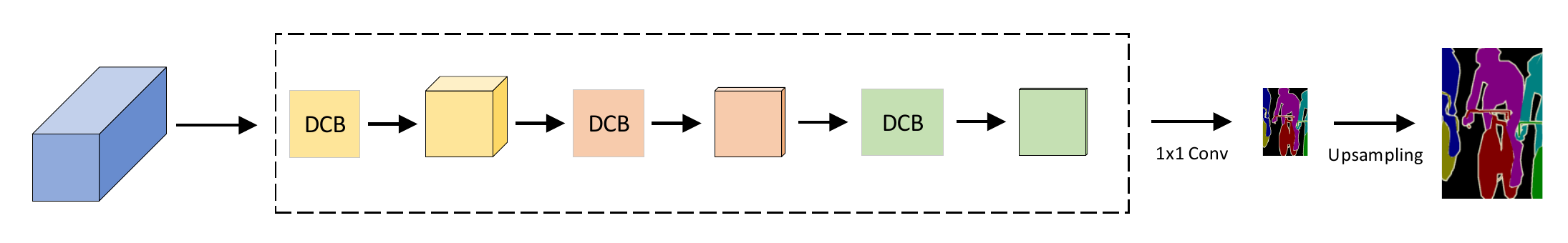}}
		\centerline{\small (b)}
	\end{minipage}
	
    \vspace{0.15cm}
	\caption{ (a) A brief illustration of context extraction based semantic segmentation pipeline. 
    (b) Proposed adaptive context aggregation  (ACE) module.}
	\label{fig:summary}
\end{figure*}
In this section, we will discuss our proposed ACE module in detail. The most relevant atrous and deformable convolution operations are first introduced, then the ACE module and the network architecture are demonstrated. 

\subsection{Convolution Operation}
Atrous convolution is chosen as the tool for the context aggregation module in ASPP~\cite{chen2017rethinking}. 
For two dimensional signals such as images, atrous convolution can be written as:
\begin{equation}
    y[i]=\sum_{k}^{K}x[i+r\cdot k]\cdot w[k],
    \label{eq:atrous}
\end{equation}
where $y$ indicates the output after atrous convolution operation, $i$ is the location, $x$ is the input signal, $r$ is the atrous rate, $w$ denotes the filter with a length of $K$ and $k$ enumerates $K$. When $r=1$, the equation stands for standard convolution. The value of $r$ controls the sampling location of atrous convolution. In ASPP, different sizes of field-of-view are obtained by setting different $r$ values. 
This observation leads to our claim that a learnable field-of-view can thus be obtained by learnable sampling location of the convolution operation. 

It is interesting to notice that the recent proposed deformable convolution~\cite{dai2017deformable,zhu2018deformable} meets our requirement where convolution with learnable sampling location is proposed.  Dai \textit{et al.} propose Deformable Convolutional Netowrks (DCNv1) in~\cite{dai2017deformable} and Zhu~\textit{et al.} propose a revised version (DCNv2) in~\cite{zhu2018deformable}. Eq.~\ref{eq:deform} presents the deformable convolution  operation from DCNv1:
\begin{equation}
y[i]=\sum_{k}^{K}x[i+k+\Delta k]\cdot w[k],
\label{eq:deform}
\end{equation}
where $\Delta k$ denotes offsets. The regular sampling locations $k$ is then augmented with the irregular offset $\Delta k$.

It is clearly observed that the main difference between Eq.~\ref{eq:atrous} and Eq.~\ref{eq:deform} is that the sampling locations of atrous convolution are always regular grid. For example, the sampling grid is square for a $3\times 3$ kernel, no matter what the value of atrous rate $r$ is. But the offset $\Delta k$  is input dependent, without regular shape constraint. Besides, compared to the manually set atrous rate $r$, the offset  $\Delta k$  is learned by the network.  In~\cite{dai2017deformable}, Zhu~\textit{et al.} further investigate deformable convolution and find out that the spatial support of the deformable convolution operation from DCNv1 can extend beyond the pertinent region. Therefore, they propose DCNv2 to let the network better focus on relevant image content by introducing a modulation mechanism to manipulate the spatial support region. This modulated deformable convolution can be expressed as follows:
\begin{equation}
y[i]=\sum_{k}^{K}x[i+k+\Delta k]\cdot w[k] \cdot \Delta m_{k},
\label{eq:deformv2}
\end{equation}
where $\Delta m_{k}$ is the learnable modulation value with range of [0,1]. This modulation value can further adjust the sampled pixel's contribution, thus the spatial support regions are adjusted better.

Therefore, in this work, we propose to employ the deformable convolution operation from DCNv2 as the tool for context aggregation.
\subsection{Network Architecture}

A brief illustration of context extraction based semantic segmentation pipeline is depicted in Figure~\ref{fig:summary} (a). For a given input image, convolution networks such as ResNet~\cite{he2016deep}, Xception~\cite{chollet2017xception}, are applied to extract feature maps, then feature aggregation is employed to extract context information. Based on the aggregated feature information, the final prediction is made and up-sampled to the original input spatial resolution. 

In this paper, we only focus on feature aggregation part. Our proposed  ACE module is shown in Figure~\ref{fig:summary} (b). After  feature extraction, the input image is represented by feature maps with size of $H \times W \times C$ where $H$ and $W$ are the height  and width of the feature maps respectively, and $C$ indicates the number of feature channels. In ACE module, three deformable convolution blocks (DCB) are applied to aggregate the feature maps. Each block is consists of ``Deformable Convolution (DConv)$\rightarrow$  BN (BatchNorm) $\rightarrow$ ReLU (Rectified Linear Unit)'' operations. The outputs size after each deformable convolution block are   $H \times W \times \frac{C}{4}$, $H \times W \times \frac{C}{8}$, and $H \times W \times \frac{C}{8}$ separately.
After ACE module, one $1\times 1$ convolution operation is applied for the final segmentation map prediction. Then the predicted result is up-sampled by `bilinear' up-sampling operation to the original image spatial resolution directly. 

\section{Experiment}
\label{sec:experiment}

In this section, we validate our proposed module on two public datasets Pascal-Context~\cite{oliva2007role} and ADE20K~\cite{zhou2017scene}. We first introduce the implementation details. Then the experiment results are presented and analyzed on these two datasets. The performance of the proposed method is evaluated in terms of two common measures, namely pixel accuracy (pixAcc) and mean Intersection of Union (mIoU).

In order to illustrate the effectiveness of the proposed method, we will compare it with ASPP and PPM. It is worth noticing that Deeplabv3 and PSPNet utilize atrous convolution for feature extraction which is memory and time consuming. In~\cite{wu2019fastfcn}, Wu \textit{et al.} propose a Joint Pyramid Up-sampling (JPU) module to replace the heavy feature extraction module. Their method (FastFCN) reduces more than three times the computation complexity and reaches slightly better performance. As a result of the limited computational resource we have, we adopt the FastFCN's feature extraction part as backbone for comparison. In other words, only the feature aggregation (\textit{head}) part is replaced with ASPP (atrous spatial pyramid pooling), PPM (pyramid pooling module) and our proposed ACE ( adaptive context encoding) module.

\subsection{Implementation Details}
The implementation is based on the PyTorch~\cite{paszke2017automatic} implementation of FastFCN~\cite{wu2019fastfcn}\footnote{https://github.com/wuhuikai/FastFCN} which is similiar to~\cite{zhang2018context}\footnote{https://github.com/zhanghang1989/PyTorch-Encoding} and the implementation of Deformable ConvNets\footnote{https://github.com/chengdazhi/Deformable-Convolution-V2-PyTorch}\textsuperscript{,}\footnote{https://github.com/open-mmlab/mmdetection}. For fair comparison, we adopt the original training strategy of FastFCN~\cite{wu2019fastfcn}. 
 Specifically, ``poly'' learning rate policy is used: $lr=baselr*(1-\frac{iter}{total\_iter})^{power}$, where $power=0.9$. The initial base learning rate is set to 0.001 for batch size 16 for PASCAL-Context~\cite{oliva2007role} and 0.01 for ADE20K~\cite{zhou2017scene}. Besides, it is adjusted relatively to batch size value if other batch size is chosen: $baselr\_adjusted=\frac{baselr}{16}*batch\_size$. The networks are trained for 80 epochs with SGD for  PASCAL-Context~\cite{oliva2007role} and 120 epochs for ADE20K~\cite{zhou2017scene}. The momentum is 0.9 and the weight decay is set to 0.0001. For data augmentation, the image is randomly flipped and scaled between 0.5 to 2. Then the image is cropped to a fixed size (480 $\times$ 480). Pixel-wise cross-entropy loss and auxiliary loss as presented in~\cite{zhao2017pyramid,wu2019fastfcn} are used, the weight for auxiliary loss is set to 0.2.

Due to limited access to multi-GPUs computational resource, our experiment includes training on a single GeForce RTX 2080 GPU for small batch sizes and  training on 4 $\times$ GeForce GTX 1080 GPUs for bath size 16. 

\subsection{Pascal-Context}
\textbf{Dataset}
Pascal-Context dataset~\cite{oliva2007role} is based on PASCAL VOC 2010 with additional annotations which provide annotations for the whole scene. Training images are 4,998 (\textit{pascal-train}) and testing images contain 5,105 images (\textit{pascal-val}). Following the prior work~\cite{chen2018deeplab,zhang2018context,wu2019fastfcn}, the semantic labels we used in this paper are the 59 categories with one background class.

\textbf{Experimental Results}
Table~\ref{tab:pcontext:val:batch} illustrates the performance on~\textit{pascal-val} of ASPP, PPM and the proposed ACE based FastFCN~\cite{wu2019fastfcn}. We trained the models on different batch sizes: 4, 6 and 16. 6 is the maximum batch size for our single GPU. The methods employ ResNet\=/50~\cite{he2016deep} as backbone. The reported results are obtained for 59 classes without multi-scale evaluation. All the methods are trained by our machine for fair comparison. Obviously, our proposed ACE outperforms ASPP and PSP on all the different batch size training settings. It's worth mentioning that,   ASPP based FastFCN's accuracy influenced severely by batch size, as also pointed out in Deeplabv3 paper~\cite{chen2017rethinking}, training DeepLabv3 model with small batch size is inefficient. Note that our proposed method not only reaches the best result, but also shows its stable performance for different batch sizes. The absolute improvements of mIoU of ACE compared to ASPP are 4.45\%, 2.83\% and 1.31\% for batch sizes 4, 6, 16 respectively.  And 2.39\%, 1.04\%, and 0.81\% compared to PPM.

\begin{table}[]
\begin{center}
\resizebox{0.9\linewidth}{!}{
\begin{tabular}{c|c|c|c}
\hline
\textbf{Batch Size}          & \textbf{Head}       &\textbf{ pixAcc\%} & \textbf{mIoU\%} \\ 
\hline
\multirow{3}{*}{4}  & ASPP     & 75.42    & 43.62  \\                     
                    & PPM      & 75.58    & 45.68  \\
                    & Proposed   & \textbf{77.68}    & \textbf{48.07}  \\
                    \hline
\multirow{3}{*}{6}  & ASPP       & 77.19    & 46.53  \\                          
                    & PPM        & 77.45    & 48.32  \\  
                    & Proposed   & \textbf{78.35}    & \textbf{49.36}  \\
                    \hline 
\multirow{3}{*}{16} & ASPP       & 78.68    & 49.04  \\
                    & PPM        & 78.41    & 49.54  \\
                    & Proposed   & \textbf{78.85}& \textbf{50.35} \\
                    \hline
                
\end{tabular}
}
\end{center}

\setlength{\belowcaptionskip}{0pt}
\setlength{\abovecaptionskip}{10pt}
\caption{Segmentation results on PASCAL-Context dataset (\textit{pascal-val}).}
\label{tab:pcontext:val:batch}
\end{table}

Table~\ref{tab:pcontext:val:sota} illustrates the performance compared to the state-of-the-art methods. For a fair comparison, the reported result of our method is calculated with background class and multi-scale evaluation where the network prediction is averaged through multiple scales as in~\cite{liu2015parsenet,zhao2017pyramid,zhang2018context,wu2019fastfcn}. The results of the other methods are obtained from the corresponding papers. Our proposed method achieves 53.6\% mIoU, which outperforms the previous methods. 

\begin{table}[ht]

\begin{center}

\resizebox{0.9\linewidth}{!}{
\begin{threeparttable}
\begin{tabular}{c|c}
\hline
\textbf{Method}         & \textbf{mIoU\%} \\ \hline
FCN-8s~\cite{long2015fully}  & 37.8            \\ 
ParseNet~\cite{liu2015parsenet}                & 40.4            \\ 
Piecewise~\cite{lin2016efficient}            & 43.3            \\ 
Deeplabv2 (Res101-COCO)~\cite{chen2018deeplab} & 45.7            \\ 
RefineNet (Res152) ~\cite{lin2017refinenet}     & 47.3            \\ 
PSPNet (Res101)~\cite{zhao2017pyramid}        & 47.8            \\ 
EncNet (Res101)~\cite{zhang2018context}        & 51.7            \\ 
DANet (Res101)~\cite{fu2018dual}        & 52.6            \\ 
FastFCN (Res101,EncNet)\tnote{*}~~\cite{wu2019fastfcn}        & 53.1            \\ \hline
Proposed (Res101)               & \textbf{53.6}   \\ \hline
\end{tabular}
\begin{tablenotes}
\raggedright
\item[*] \footnotesize FastFCN backbone with EncNet \textit{head}.
\end{tablenotes}

\end{threeparttable}
}
\end{center}

\setlength{\belowcaptionskip}{0pt}
\setlength{\abovecaptionskip}{10pt}
\caption{Segmentation results on PASCAL-Context dataset (\textit{pascal-val}) of the state-of-the-art methods.}
\label{tab:pcontext:val:sota}
\end{table}





\subsection{ADE20K}

\textbf{Dataset}
ADE20K  is used in ImageNet Scene parsing challenge 2016 and it contains 150 object categories. It is divided into 20k/2K/3K images for training (\textit{a-train}), validation (\textit{a-val}) and testing (\textit{a-test}) respectively.

\textbf{Experimental Results}
Table~\ref{tab:ade20k:val} demonstrates the results for \textit{a-val} set without multi-scale evaluation. Note, one GeForce RTX 2080 GPU maximum can fit 4 batches, thus the results reported in Table~\ref{tab:ade20k:val} are obtained on batch size 4 and ResNet\=/50 as backbone. 
Our proposed method achieves the best result compared to ASPP and PPM based FastFCN with an absolute improvement of 1.4\% and 0.71\% for ASPP and PPM in terms of mIoU.

\begin{table}[htb]
\begin{center}
\resizebox{1\linewidth}{!}{
\begin{tabular}{c|c|c|c}
	\hline
	Batch Size         & \textbf{Head} & \textbf{pixAcc\%} &\textbf{mIoU\%} \\ \hline 
	\multirow{3}{*}{4} & ASPP                           & 78.11                              & 37.11                            \\ 
	& PPM                            & 77.39                              & 37.80                            \\
	& Proposed                       & \textbf{78.62}    & \textbf{38.51}  \\ \hline
\end{tabular}
}
\end{center}

\setlength{\belowcaptionskip}{0pt}
\setlength{\abovecaptionskip}{10pt}
\caption{Segmentation results on ADE20K dataset (\textit{a-val}).}
\label{tab:ade20k:val}
\end{table}

In order to compare with the state-of-the-art methods, we further train our model with ResNet\=/101 backbone on 4$\times$GeFore 1080 GPUs with batch size 16. Table~\ref{tab:ade20k:val_sota} shows the obtained result and results reported in the corresponding papers of the  other approaches. The proposed method provides better result compared to PSPNet with an absolute improvement of 0.52\% of mIoU. EncNet achieves the best result. Except the methodology itself, some part of the performance gap could be from the training strategy, such as EncNet is trained with image size of 576$\times$576 and our method is trained with 480$\times$480. 

\begin{table}[ht]
	\begin{center}
		\resizebox{1\linewidth}{!}{
\begin{threeparttable}		
\begin{tabular}{ccc}
	\hline
	\textbf{Method}                       & \textbf{pixAcc\%} & \textbf{mIoU\%}                    \\ \hline
	FCN~\cite{long2015fully}                                   & 71.32             & 29.39                              \\
	SegNet~\cite{badrinarayanan2017segnet}                                & 71.00             & 21.64                              \\
	DilatedNet~\cite{YuKoltun2016}                            & 73.55             & 32.31                              \\
	CascadeNet~\cite{zhou2017scene}                            & 74.52             & 34.90                              \\
	RefineNet (Res152)~\cite{lin2017refinenet}                    & -                 & 40.7                               \\
	PSPNet (Res101)~\cite{zhao2017pyramid}  & 81.39             & 43.29                              \\
	EncNet (Res101)~\cite{zhang2018context}                       & \textbf{81.69}    & \textbf{44.65}                     \\
	FastFCN (Res101,EncNet)\tnote{*}~~\cite{wu2019fastfcn}  & 80.99             & 44.34                              \\\hline
	Proposed                              & 81.07             & 43.81                  				\\
	\hline           
\end{tabular}
\begin{tablenotes}
\raggedright
\item[*] \footnotesize FastFCN backbone with EncNet \textit{head}.
\end{tablenotes}

\end{threeparttable}
}
\end{center}
\setlength{\belowcaptionskip}{0pt}
\setlength{\abovecaptionskip}{10pt}
\caption{Segmentation results on ADE20K dataset (\textit{a-val}) of  the state-of-the-art methods. }
\label{tab:ade20k:val_sota}
\end{table}
 Moreover, we fine-tune our trained model for another 20 epochs on \textit{a-train} \textit{a-val} set with a smaller learning rate 0.001, then submit the \textit{a-test} set result to the evaluation website~\footnote{http://sceneparsing.csail.mit.edu/}. Our method obtains 72.99\% (pixAcc) and 37.71\% (mIoU) with a final score of 0.5535 which is not the best but is an encouraging result.


\section{Discussion and Conclusion}
\label{sec:conclusion}
In summary, in this work, we revisited the atrous convolution operation and pyramid pooling modules and propose an  effective feature aggregation method based on deformable convolution to adaptively extract multiple scale context for the final segmentation map prediction. Based on the experimental validation, our method outperforms the ASPP module and PPM on Pascal-Context and ADE20K datasets. Noticeably, although our goal for this work is to propose a better multiple scale context aggregation module, rather than to obtain the best results on the benchmarks, our proposed approach achieves state-of-the-art result 53.6\% mIOU on Pascal-Conext and encouraging result 0.5535 on ADE20K.

 All the experiments confirm that an adaptive context encoding (ACE) module is benefit for semantic segmentation which deserve further research. In this work, we directly use deformable convolution as the tool for ACE and simply cascaded three deformable convolution blocks, a sophisticated designed architecture is essential. We believe that further exploration for the usage and improvement of our feature aggregation idea is promising  and necessary in the design of an efficient semantic segmentation.


{\small
\bibliographystyle{ieee}
\bibliography{egbib}

\begin{thebibliography}{10}\itemsep=-1pt

\bibitem{amirul2017gated}
M.~Amirul~Islam, M.~Rochan, N.~D. Bruce, and Y.~Wang.
\newblock Gated feedback refinement network for dense image labeling.
\newblock In {\em Proceedings of the IEEE Conference on Computer Vision and
  Pattern Recognition}, pages 3751--3759. IEEE, 2017.

\bibitem{badrinarayanan2017segnet}
V.~Badrinarayanan, A.~Kendall, and R.~Cipolla.
\newblock Segnet: A deep convolutional encoder-decoder architecture for image
  segmentation.
\newblock {\em IEEE transactions on pattern analysis and machine intelligence},
  39(12):2481--2495, 2017.

\bibitem{chen2014semantic}
L.-C. Chen, G.~Papandreou, I.~Kokkinos, K.~Murphy, and A.~L. Yuille.
\newblock Semantic image segmentation with deep convolutional nets and fully
  connected crfs.
\newblock In {\em International Conference on Learning Representations}, 2015.

\bibitem{chen2018deeplab}
L.-C. Chen, G.~Papandreou, I.~Kokkinos, K.~Murphy, and A.~L. Yuille.
\newblock Deeplab: Semantic image segmentation with deep convolutional nets,
  atrous convolution, and fully connected crfs.
\newblock {\em IEEE transactions on pattern analysis and machine intelligence},
  40(4):834--848, 2018.

\bibitem{chen2017rethinking}
L.-C. Chen, G.~Papandreou, F.~Schroff, and H.~Adam.
\newblock Rethinking atrous convolution for semantic image segmentation.
\newblock {\em arXiv preprint arXiv:1706.05587}, 2017.

\bibitem{chen2018encoder}
L.-C. Chen, Y.~Zhu, G.~Papandreou, F.~Schroff, and H.~Adam.
\newblock Encoder-decoder with atrous separable convolution for semantic image
  segmentation.
\newblock In {\em Proceedings of the European Conference on Computer Vision},
  pages 801--818. Springer, 2018.

\bibitem{chollet2017xception}
F.~Chollet.
\newblock Xception: Deep learning with depthwise separable convolutions.
\newblock In {\em Proceedings of the IEEE Conference on Computer Vision and
  Pattern Recognition}, pages 1251--1258. IEEE, 2017.

\bibitem{dai2017deformable}
J.~Dai, H.~Qi, Y.~Xiong, Y.~Li, G.~Zhang, H.~Hu, and Y.~Wei.
\newblock Deformable convolutional networks.
\newblock In {\em Proceedings of the IEEE International Conference on Computer
  Vision}, pages 764--773. IEEE, 2017.

\bibitem{farabet2013learning}
C.~Farabet, C.~Couprie, L.~Najman, and Y.~LeCun.
\newblock Learning hierarchical features for scene labeling.
\newblock {\em IEEE transactions on pattern analysis and machine intelligence},
  35(8):1915--1929, 2013.

\bibitem{gu2019net}
Z.~Gu, J.~Cheng, H.~Fu, K.~Zhou, H.~Hao, Y.~Zhao, T.~Zhang, S.~Gao, and J.~Liu.
\newblock Ce-net: Context encoder network for 2d medical image segmentation.
\newblock {\em IEEE transactions on medical imaging}, 2019.

\bibitem{he2016deep}
K.~He, X.~Zhang, S.~Ren, and J.~Sun.
\newblock Deep residual learning for image recognition.
\newblock In {\em Proceedings of the IEEE Conference on Computer Vision and
  Pattern Recognition}, pages 770--778. IEEE, 2016.

\bibitem{fu2018dual}
H.~T. Y. L. Y. B. Z. F. a. H.~L. Jun~Fu, Jing~Liu.
\newblock Dual attention network for scene segmentation.
\newblock In {\em Proceedings of the IEEE Conference on Computer Vision and
  Pattern Recognition}. IEEE, 2019.

\bibitem{lin2017refinenet}
G.~Lin, A.~Milan, C.~Shen, and I.~Reid.
\newblock Refinenet: Multi-path refinement networks for high-resolution
  semantic segmentation.
\newblock In {\em Proceedings of the IEEE Conference on Computer Vision and
  Pattern Recognition}, pages 1925--1934. IEEE, 2017.

\bibitem{lin2016efficient}
G.~Lin, C.~Shen, A.~Van Den~Hengel, and I.~Reid.
\newblock Efficient piecewise training of deep structured models for semantic
  segmentation.
\newblock In {\em Proceedings of the IEEE Conference on Computer Vision and
  Pattern Recognition}, pages 3194--3203. IEEE, 2016.

\bibitem{liu2015parsenet}
W.~Liu, A.~Rabinovich, and A.~C. Berg.
\newblock Parsenet: Looking wider to see better.
\newblock {\em arXiv preprint arXiv:1506.04579}, 2015.

\bibitem{long2015fully}
J.~Long, E.~Shelhamer, and T.~Darrell.
\newblock Fully convolutional networks for semantic segmentation.
\newblock In {\em Proceedings of the IEEE Conference on Computer Vision and
  Pattern Recognition}, pages 3431--3440. IEEE, 2015.

\bibitem{mallat1999wavelet}
S.~Mallat.
\newblock {\em A wavelet tour of signal processing}.
\newblock Elsevier, 1999.

\bibitem{mohammed2018net}
A.~Mohammed, S.~Yildirim, I.~Farup, M.~Pedersen, and {\O}.~Hovde.
\newblock Y-net: A deep convolutional neural network for polyp detection.
\newblock {\em arXiv preprint arXiv:1806.01907}, 2018.

\bibitem{mohammed2019streoscennet}
A.~Mohammed, S.~Yildirim, I.~Farup, M.~Pedersen, and {\O}.~Hovde.
\newblock Streoscennet: surgical stereo robotic scene segmentation.
\newblock In {\em Medical Imaging 2019: Image-Guided Procedures, Robotic
  Interventions, and Modeling}, volume 10951, page 109510P. International
  Society for Optics and Photonics, 2019.

\bibitem{noh2015learning}
H.~Noh, S.~Hong, and B.~Han.
\newblock Learning deconvolution network for semantic segmentation.
\newblock In {\em Proceedings of the IEEE International Conference on Computer
  Vision}, pages 1520--1528. IEEE, 2015.

\bibitem{oliva2007role}
A.~Oliva and A.~Torralba.
\newblock The role of context in object recognition.
\newblock {\em Trends in cognitive sciences}, 11(12):520--527, 2007.

\bibitem{paszke2017automatic}
A.~Paszke, S.~Gross, S.~Chintala, G.~Chanan, E.~Yang, Z.~DeVito, Z.~Lin,
  A.~Desmaison, L.~Antiga, and A.~Lerer.
\newblock Automatic differentiation in pytorch.
\newblock 2017.

\bibitem{peng2017large}
C.~Peng, X.~Zhang, G.~Yu, G.~Luo, and J.~Sun.
\newblock Large kernel matters—improve semantic segmentation by global
  convolutional network.
\newblock In {\em Proceedings of the IEEE Conference on Computer Vision and
  Pattern Recognition}, pages 1743--1751. IEEE, 2017.

\bibitem{pinheiro2014recurrent}
P.~H. Pinheiro and R.~Collobert.
\newblock Recurrent convolutional neural networks for scene labeling.
\newblock In {\em International Conference on Machine Learning}, 2014.

\bibitem{ronneberger2015u}
O.~Ronneberger, P.~Fischer, and T.~Brox.
\newblock U-net: Convolutional networks for biomedical image segmentation.
\newblock In {\em International Conference on Medical Image Computing and
  Computer-assisted Intervention}, pages 234--241. Springer, 2015.

\bibitem{vaswani2017attention}
A.~Vaswani, N.~Shazeer, N.~Parmar, J.~Uszkoreit, L.~Jones, A.~N. Gomez,
  {\L}.~Kaiser, and I.~Polosukhin.
\newblock Attention is all you need.
\newblock In {\em Advances in neural information processing systems}, pages
  5998--6008, 2017.

\bibitem{wu2019fastfcn}
H.~Wu, J.~Zhang, K.~Huang, K.~Liang, and Y.~Yu.
\newblock Fastfcn: Rethinking dilated convolution in the backbone for semantic
  segmentation.
\newblock {\em arXiv preprint arXiv:1903.11816}, 2019.

\bibitem{yu2018bisenet}
C.~Yu, J.~Wang, C.~Peng, C.~Gao, G.~Yu, and N.~Sang.
\newblock Bisenet: Bilateral segmentation network for real-time semantic
  segmentation.
\newblock In {\em Proceedings of the European Conference on Computer Vision},
  pages 325--341. Springer, 2018.

\bibitem{yu2018learning}
C.~Yu, J.~Wang, C.~Peng, C.~Gao, G.~Yu, and N.~Sang.
\newblock Learning a discriminative feature network for semantic segmentation.
\newblock In {\em Proceedings of the IEEE Conference on Computer Vision and
  Pattern Recognition}, pages 1857--1866. IEEE, 2018.

\bibitem{YuKoltun2016}
F.~Yu and V.~Koltun.
\newblock Multi-scale context aggregation by dilated convolutions.
\newblock In {\em International Conference on Learning Representations}, 2016.

\bibitem{yuan2018ocnet}
Y.~Yuan and J.~Wang.
\newblock Ocnet: Object context network for scene parsing.
\newblock {\em arXiv preprint arXiv:1809.00916}, 2018.

\bibitem{zeiler2014visualizing}
M.~D. Zeiler and R.~Fergus.
\newblock Visualizing and understanding convolutional networks.
\newblock In {\em Proceedings of the European Conference on Computer Vision},
  pages 818--833. Springer, 2014.

\bibitem{zhang2018context}
H.~Zhang, K.~Dana, J.~Shi, Z.~Zhang, X.~Wang, A.~Tyagi, and A.~Agrawal.
\newblock Context encoding for semantic segmentation.
\newblock In {\em Proceedings of the IEEE Conference on Computer Vision and
  Pattern Recognition}, pages 7151--7160. IEEE, 2018.

\bibitem{zhao2017pyramid}
H.~Zhao, J.~Shi, X.~Qi, X.~Wang, and J.~Jia.
\newblock Pyramid scene parsing network.
\newblock In {\em Proceedings of the IEEE Conference on Computer Vision and
  Pattern Recognition}, pages 2881--2890. IEEE, 2017.

\bibitem{zhou2017scene}
B.~Zhou, H.~Zhao, X.~Puig, S.~Fidler, A.~Barriuso, and A.~Torralba.
\newblock Scene parsing through ade20k dataset.
\newblock In {\em Proceedings of the IEEE Conference on Computer Vision and
  Pattern Recognition}, pages 633--641. IEEE, 2017.

\bibitem{zhu2018deformable}
X.~Zhu, H.~Hu, S.~Lin, and J.~Dai.
\newblock Deformable convnets v2: More deformable, better results.
\newblock {\em arXiv preprint arXiv:1811.11168}, 2018.

\end{thebibliography}
}

\end{document}